\documentclass[runningheads]{llncs}
\usepackage{graphicx}
\usepackage{hyperref}
\usepackage{color}

\usepackage[binary-units=true,detect-weight=true]{siunitx}

\usepackage{amsmath}
\usepackage{amsfonts}

\usepackage{cite}

\usepackage{todonotes}

\newcommand{\spass}{SPASS}
\newcommand{\vampire}{Vampire}
\newcommand{\avatar}{AVATAR}
\newcommand{\smtlib}{SMT-LIB}

\begin{document}
%
\title{Improving ENIGMA-Style Clause Selection While Learning From History}
%
%
\author{Martin Suda
\orcidID{0000-0003-0989-5800}}
\authorrunning{M. Suda}
%
\institute{Czech Technical University in Prague, Czech Republic \\
\email{martin.suda@cvut.cz}}
\maketitle              
\begin{abstract}
We re-examine the topic of machine-learned clause selection guidance in saturation-based theorem provers.
The central idea, recently popularized by the ENIGMA system,
 is to learn a classifier for recognizing 
clauses that appeared in previously discovered proofs. 
In subsequent runs, clauses classified positively are prioritized for selection.
We propose several improvements to this approach and experimentally confirm their viability. 
%
%
%
For the demonstration, we use a recursive neural network 
to classify clauses based on their derivation history and the presence or absence of automatically supplied theory axioms therein.
The automatic theorem prover \vampire{} guided by the network
achieves a \SI{41}{\percent} improvement on a relevant subset of \smtlib{} in a real time evaluation.

\keywords{Saturation-based theorem proving \and Clause Selection \and Machine Learning \and Recursive Neural Networks.}
\end{abstract}


\section{Introduction}


The idea to improve the performance of saturation-based automatic theorem provers (ATPs)
with the help of machine learning (ML), while going back at least to the early work of Schulz \cite{DS1996b,Schulz:Diss-2000},
has recently been enjoying a renewed interest.
Most notable is the ENIGMA system \cite{DBLP:conf/mkm/JakubuvU17,DBLP:conf/mkm/JakubuvU18}
extending the ATP {\sc E}~\cite{SCV:CADE-2019} by machine learned clause selection guidance.
The architecture trains a binary classifier for recognizing as positive those 
clauses that appeared in previously discovered proofs and as negative the remaining selected ones.
In subsequent runs, clauses classified positively are prioritized for selection.

A system such as ENIGMA needs to carefully balance the expressive power of the used ML model
with the time it takes to evaluate its advice.
For example, Loos et al.~\cite{DBLP:conf/lpar/LoosISK17}, who were the first
to integrate state-of-the-art neural networks with {\sc E}, discovered their models to be too slow
to simply replace the traditional clause selection mechanism.
In the meantime, the data-hungry deep learning approaches motivate researchers to augment 
training data with artificially crafted theorems \cite{DBLP:journals/corr/abs-2006-11259}.
Yet another interesting aspect is what features we allow the model to learn from.
One could speculate that the recent success of ENIGMA on the Mizar dataset 
\cite{DBLP:conf/cade/ChvalovskyJ0U19,DBLP:conf/itp/JakubuvU19} 
can at least partially be explained by the involved problems 
sharing a common source and encoding. It is still open whether 
some new form of general ``theorem proving knowledge'' could be learned
to improve the performance of an ATP across, e.g., the very diverse TPTP library. 

In this paper, we propose several improvements to ENIGMA-style clause selection guidance
and experimentally test their viability in a novel setting:
\begin{itemize}
	\item
	We lay out a set of possibilities for integrating the learned advice into the ATP
	and single out the recently developed layered clause selection \cite{DBLP:conf/ijcai/Gleiss020,DBLP:conf/cade/Gleiss020,DBLP:conf/cade/Tammet19} 
	as particularly suitable for the task.

	\item We speed up evaluation 
	by a new lazy evaluation scheme under 
	which many generated clauses need not be 
	evaluated by the potentially slow classifier. 

	\item
	We demonstrate the importance of ``positive bias'',
	i.e., of tuning the classifier to rather err on the side of false 
	positives than on the side of false negatives.
	
	\item
	Finally, we propose the use of ``negative mining'' 
	for improving learning from proofs 
	obtained while relying on previously learned guidance.	
\end{itemize}
To test these ideas, we designed a recursive neural network 
to classify clauses based solely on their derivation history
and the presence or absence of automatically supplied theory axioms therein.
This allows us to test here, as a byproduct of the conducted experiments, 
whether the human-engineered heuristic 
for controlling the amount of theory reasoning presented in our previous work\cite{DBLP:conf/cade/Gleiss020}
can be matched or even overcome by the automatically discovered neural guidance.

The rest of the paper is structured as follows. Sect.~\ref{sec:guidance}
recalls the necessary ATP theory, explains clause selection and how to improve it using ML. 
Sect.~\ref{sec:layered} covers layered clause selection and the new lazy evaluation scheme.
In Sect.~\ref{sec:neural_training}, we describe our neural architecture 
and in Sect.~\ref{sec:experiments} we bring everything together and evaluate
the presented ideas, using the prover \vampire{} as our workhorse and a 
relevant subset of \smtlib{} as the testing grounds. 
Finally, Sect.~\ref{sec:conclusion} concludes.

\section{ATPs, Clause Selection, and Machine Learning} \label{sec:guidance}

The technology behind the
modern automatic theorem provers (ATPs) for first-order logic (FOL),
such as {\sc E}~\cite{SCV:CADE-2019}, \spass{} \cite{DBLP:conf/cade/WeidenbachDFKSW09}, or \vampire{}~\cite{DBLP:conf/cav/KovacsV13},
can be roughly outlined by using the following three adjectives.

\paragraph{Refutational:} The task of the prover is to 
check whether a given conjecture $G$ logically follows from given axioms $A_1,\ldots,A_n$, i.e. whether
\begin{equation} \label{eq:entails}
A_1,\ldots,A_n \models G, 
\end{equation}
where $G$ and each $A_i$ are FOL formulas. The prover starts by negating the conjecture $G$ and
transforming $\neg G, A_1,\ldots,A_n$ into an equisatisfiable set of clauses $\mathcal{C}$.
It then applies a sound logical calculus to iteratively derive further clauses, 
logical consequence of $\mathcal{C}$, until the obvious contradiction in the form of the empty clause $\bot$ is derived.
This \emph{refutes} the assumption that $\neg G, A_1,\ldots,A_n$ could be satisfiable and thus confirms \eqref{eq:entails}.

\paragraph{Superposition-based:}
The most popular calculus used in this context is superposition \cite{DBLP:journals/logcom/BachmairG94,DBLP:books/el/RV01/NieuwenhuisR01},
an extension of ordered resolution \cite{DBLP:books/el/RV01/BachmairG01} with a built-in support 
for handling equality.
It consists of several inference rules, such as the resolution rule, factoring, subsumption, superposition, or demodulation.

Inference rules in general determine how to derive new clauses from old ones, where by old clauses we mean either the 
initial clauses $\mathcal{C}$ or clauses derived previously. The clauses that need to be present for a rule to be applicable
are called the \emph{premises} and the newly derived clause is called the \emph{conclusion}.
By applying the inference rules the prover gradually constructs a \emph{derivation}, a directed acyclic (hyper-)graph (DAG),
with the initial clauses forming the leaves and the derived clauses (labeled by the respective applied rules) forming the internal nodes.
A \emph{proof} is the smallest sub-DAG of a derivation containing the final empty clause
and for every derived clause the corresponding inference and its premises.


\paragraph{Saturation-based:} A saturation algorithm is the concrete way of organizing the process of deriving new clauses,
such that every applicable inference is eventually considered. Modern saturation-based ATPs employ some variant of 
the \emph{given-clause algorithm}, in which clauses are selected for inferences one by one \cite{DBLP:journals/jsc/RiazanovV03}.

The process employs two sets of clauses, often called the \emph{active} set $\mathcal{A}$ and the \emph{passive} set $\mathcal{P}$.
At the beginning all the initial clauses are put to the passive set.
Then in every iteration, the prover \emph{selects} and removes a clause $C$ from $\mathcal{P}$,
inserts it into $\mathcal{A}$, and performs all the applicable inferences with premises in $\mathcal{A}$ 
such that at least one of the premises is $C$. The conclusions of these inferences are then inserted into $\mathcal{P}$.
This way the prover maintains (at the end of each iteration) the invariant that inferences among the clauses in the active set have been performed. 
The selected clause $C$ is sometimes also called the ``given clause''.

During a typical prover run, $\mathcal{P}$ grows much faster than $\mathcal{A}$ (the growth is roughly quadratic).
Analogously, although for different reasons, when a proof is discovered, 
its clauses constitute only a fraction of $\mathcal{A}$. 
Notice that every clause $C \in \mathcal{A}$ that is in the end \emph{not} part of the proof 
did not need to be selected and represents a wasted effort. This explains
why \emph{clause selection}, i.e.~the procedure for picking in each iteration the next clause to process,
is one of the main heuristic decision points in the prover, 
which hugely affects its performance \cite{DBLP:conf/cade/SchulzM16}.


\subsection{Traditional Approaches to Clause Selection}

There are two basic criteria that have been identified as generally correlating 
with the likelihood of a clause contributing to the yet-to-be discovered proof.

One is clause's \emph{age} or, more precisely, its ``date of birth'',
typically implemented as an ever increasing timestamp. 
Preferring for selection old clauses to more recently derived ones
corresponds to a breadth-first strategy and ensures fairness.
The other criterion is clause's size, referred to as \emph{weight} in the ATP lingo,
and is realized by some form of symbol counting.
Preferring for selection small clauses to large ones is a greedy strategy,
based on the observation that small conclusions typically belong to inferences with small premises
and that the ultimate conclusion---the empty clause---is the smallest of all.\footnote{
As pointed out by an anonymous reviewer, 
in the presence of subsumption, all symbol-counting schemes
that assign consistent positive weights to each symbol are fair in
themselves (as modulo variable renaming, there is only a finite
number of clauses with a weight below a given threshold).}
The best results are achieved when these two criteria (or their variations) are combined \cite{DBLP:conf/cade/SchulzM16}.
 
To implement efficient clause selection by numerical criteria such as age and weight,
an ATP represents the passive set $\mathcal{P}$ as a set of priority queues.
A queue contains (pointers to) the clauses in $\mathcal{P}$ ordered by its respective criterion.
Selection typically alternates between the available queues under a certain ratio. 
A successful strategy is, for instance, to select 10 clauses by weight for every clause selected by age, i.e.,
with an \emph{age-to-weight} ratio of 1:10.




\subsection{ENIGMA-style Machine-Learned Clause Selection Guidance}

The idea to improve clause selection by learning from previous prover
experience goes, to the best of our knowledge, back to Schulz \cite{Schulz:Diss-2000,DS1996b}
and has more recently been successfully employed by the ENIGMA system and others 
\cite{DBLP:conf/mkm/JakubuvU17,DBLP:conf/mkm/JakubuvU18,DBLP:conf/cade/ChvalovskyJ0U19,DBLP:conf/cade/JakubuvCOP0U20,DBLP:conf/lpar/LoosISK17}.

The experience is collected from successful prover runs, where each selected clause constitutes a training example 
and the example is marked as \emph{positive}, if the clause ended-up in the discovered proof, and \emph{negative} otherwise.
A machine learning (ML) algorithm is then used to \emph{fit} this data and produce a \emph{model} $\mathcal{M}$
for \emph{classifying} clauses into positive and negative, accordingly.
A good learning algorithm produces a model $\mathcal{M}$ which not only accurately
classifies the training data but also \emph{generalizes} well to unseen examples.
The computational costs of both training and evaluation are also important.

While clauses are logical formulas, i.e., discrete objects forming a countable set,
ML algorithms, rooted in mathematical statistics, 
are primarily equipped to dealing with fixed-seized real-valued vectors.
Thus the question of how to \emph{represent} clauses for the learning
is the first obstacle that needs to be overcome, before the whole idea can be made to work.
In the beginning, the authors of ENIGMA experimented with various forms of hand-crafted numerical clause \emph{features}
\cite{DBLP:conf/mkm/JakubuvU17,DBLP:conf/mkm/JakubuvU18}. An attractive alternative 
explored in later work \cite{DBLP:conf/lpar/LoosISK17,DBLP:conf/cade/ChvalovskyJ0U19,DBLP:conf/cade/JakubuvCOP0U20}
is the use of artificial \emph{neural networks}, which can be understood as extracting the most relevant features automatically.

An important distinction can in both cases be made between approaches
which have access to the concrete identity of predicate and function symbols (i.e., the signature)
that make up the clauses, and those that do not. For example: Is the ML algorithm
allowed to assume that the symbol \texttt{grp\_mult} is used to represent the multiplication operation in a group 
or does it only recognize a general binary function? The first option can be much more powerful,
but we need to ensure that the signature symbols are \emph{aligned} and 
used consistently across the problems in our benchmark. Otherwise the learned advice 
cannot meaningfully cary over to previously unsolved problems. 
While the assumption of aligned signature has been employed by the early systems \cite{DBLP:conf/mkm/JakubuvU17,DBLP:conf/lpar/LoosISK17},
the most recent version of ENIGMA \cite{DBLP:conf/cade/JakubuvCOP0U20,DBLP:conf/ecai/OlsakKU20} can work in a ``signature agnostic'' mode. 

In this work we represent clauses solely by their derivation history, deliberately ignoring their logical content.
Thus we do not require the assumption of an aligned signature, per se. However, we rely on a 
fixed set of distinguished axioms to supply features in the derivation leaves.


\subsection{Integrating the Learned Advice} \label{sec:integration}


Once we have a trained model $\mathcal{M}$, an immediate possibility for integrating it 
into the clause selection procedure is to introduce a new queue that will order the clauses using $\mathcal{M}$. 
Two basic versions of this idea have been described:

\paragraph{``Priority'':} The ordering puts all the clauses classified by $\mathcal{M}$ as positive before 
those classified negatively. Within the two classes, older clauses are preferred.

Let us for the purposes of future reference denote this scheme $\mathcal{M}^{1,0}$.
It has been successfully used by the early ENIGMAs \cite{DBLP:conf/mkm/JakubuvU17,DBLP:conf/mkm/JakubuvU18,DBLP:conf/cade/ChvalovskyJ0U19}.

\paragraph{``Logits'':} Even models officially described as binary classifiers
typically internally compute a real-valued estimate $L$ of how much ``positive'' or ``negative'' an example
appears to be and only turn this estimate into a binary decision in the last step,
by comparing it against a fixed threshold $t$, most often 0. A machine learning term for this estimate $L$ is the \emph{logit}.\footnote{
A logit can be turned into a (formal) probability, i.e. a value between 0 and 1, by passing it, as is typically done,
through the \emph{sigmoid} function $\sigma(x) = 1/(1+e^{-x})$.}

The second version orders the clauses on the new queue by the ``raw'' logits produced by a model.
We denote it $\mathcal{M}^{\mathbb{-R}}$ to stress that clauses with high $L$ are treated as small
from the perspective of the selection and therefore preferred.
This scheme has been used by Loos et al.~\cite{DBLP:conf/lpar/LoosISK17} and in the latest ENGIMA 
\cite{DBLP:conf/cade/JakubuvCOP0U20,logits_in_mireks_network}.

\paragraph{Combining with a traditional strategy.}
While it is possible to rely exclusively on selection governed by the model,
it turns out to be better \cite{DBLP:conf/cade/ChvalovskyJ0U19}
to combine it with the traditional heuristics. The most natural choice is to take $\mathcal{S}$, 
the original strategy that was used to generate the training data,
and extend it by adding the new queue, be it $\mathcal{M}^{1,0}$ or $\mathcal{M}^{\mathbb{-R}}$,
next to the already present queues. We then again supply a ratio 
under which the original selection from $\mathcal{S}$ and the new selection based on $\mathcal{M}$ get alternated.
We will denote this kind of combination with the original strategy as 
$\mathcal{S}\oplus \mathcal{M}^{1,0}$ and
$\mathcal{S}\oplus \mathcal{M}^{\mathbb{-R}}$, respectively.

\section{Layered Clause Selection and Lazy Model Evaluation} \label{sec:layered}

Layered clause selection (LCS) is a recently developed method \cite{DBLP:conf/ijcai/Gleiss020,DBLP:conf/cade/Gleiss020,DBLP:conf/cade/Tammet19}
for smoothly incorporating a categorical preference for certain clauses into a base clause selection strategy $\mathcal{S}$.
In this paper, we will readily use it in combination with the binary classifier advice from a trained model $\mathcal{M}$.

When we instantiate LCS to our particular case,\footnote{
We rely here on the \emph{monotone} mode of split; there is also a \emph{disjoint} mode \cite{DBLP:conf/ijcai/Gleiss020}.} 
its function can be summarized by the expression \[\mathcal{S} \oplus \mathcal{S}[\mathcal{M}^{1}].\]
In words, the base selection strategy $\mathcal{S}$ is alternated with $\mathcal{S}[\mathcal{M}^{1}]$, 
the same selection scheme $\mathcal{S}$ but applied only to clauses classified positively by $\mathcal{M}$. Implicit here is a convention
that whenever there is no positively classified passive clause,
a fallback to plain $\mathcal{S}$ occurs. Additionally, we again specify a ``second-level'' ratio
to govern the alternation between pure $\mathcal{S}$ and $\mathcal{S}[\mathcal{M}^{1}]$.

The main advantage of LCS, compared to the options outlined in the previous section, 
is that the original, typically well-tuned, base selection mechanism $\mathcal{S}$ is 
also applied to $\mathcal{M}^{1}$, the clauses classified positively by $\mathcal{M}$.

\subsection{Lazy Model Evaluation} \label{sec:lazy}

It is often the case that evaluating a clause by the model $\mathcal{M}$ is a relatively expensive operation \cite{DBLP:conf/lpar/LoosISK17}. As we explain here, however, this operation can be avoided in many cases, especially when using LCS to integrate the advice. 

We propose the following \emph{lazy evaluation approach} to be used with $\mathcal{S} \oplus \mathcal{S}[\mathcal{M}^{1}]$.
Every clause entering the passive set $\mathcal{P}$ is initially inserted 
to both $\mathcal{S}$ and $\mathcal{S}[\mathcal{M}^{1}]$
\emph{without} being evaluated by $\mathcal{M}$. Then, whenever (as governed by the second-level ratio) it 
is the moment to select a clause from $\mathcal{S}[\mathcal{M}^{1}]$, the algorithm
\begin{enumerate}
\item \label{item:lazy_eval_selection}
picks (as usual, according to $\mathcal{S}$) the best clause $C$ in $\mathcal{S}[\mathcal{M}^{1}]$,
\item
only then evaluates $C$ by $\mathcal{M}$, and
\item
if $C$ gets classified as negative, it forgets $C$, a goes back to \ref{item:lazy_eval_selection}.
\end{enumerate}
This repeats until the first positively classified clause is found,
which is then returned. Note that this way the ``observable behaviour'' of $\mathcal{S}[\mathcal{M}^{1}]$ is preserved.

The power of lazy evaluation lies in the fact that not every clause needs to be evaluated before a proof is found.
Indeed, recall the remark that the passive set $\mathcal{P}$ is typically much larger than the active set $\mathcal{A}$,
which also holds on a typical successful termination. Every clause left in passive at that moment 
is a clause that did not need to be evaluated by $\mathcal{M}$ thanks to lazy evaluation.

We remark that lazy evaluation can similarly be used with the integration mode $\mathcal{M}^{1,0}$ based on priorities.

We experimentally demonstrate the effect of the technique in Sect.~\ref{sec:lazy_exper}.


\section{A Neural Classification of Clause Derivations} \label{sec:neural_training}

In this work we choose to represent a clause, for the purpose of learning, 
solely by its derivation history.
Thus a clause can only be distinguished by the axioms from which it was derived 
and by the precise way in which these axioms interacted with each other through inferences
in the derivation.
This means we deliberately ignore the clause's logical content.

We decided to focus on this representation, because it promises to be fast.
Although an individual clause's derivation history may be large,
it is a simple function of its parents' histories (just one application of an inference rule).
Moreover, before a clause with a complicated history can be selected,
most of its ancestors will have been selected already.\footnote{
Exceptions are caused by simplifying inferences applied eagerly outside of the governance of the main clause selection mechanism.}
This guarantees the amortised cost of evaluating a single clause to be constant.

A second motivation comes from our recent work \cite{DBLP:conf/cade/Gleiss020},
where we have shown that theory reasoning facilitated by 
automatically adding theory axioms for axiomatising theories,
while in itself a powerful technique, often leads the prover to unpromising parts of the search space.
We developed a heuristic for controlling the amount of 
theory reasoning in the derivation of a clause \cite{DBLP:conf/cade/Gleiss020}.
Our goal here is to test whether a similar or even stronger heuristic can be automatically discovered 
by a neural network.

Examples of axioms that \vampire{} uses to axiomatise theories include 
the commutativity or associativity axioms for the arithmetic operations, 
an axiomatization of the theory of arrays \cite{DBLP:conf/vmcai/BradleyMS06}
or of the theory of term algebras \cite{DBLP:conf/popl/KovacsRV17}.
For us it is mainly important that the axioms are introduced internally by the prover
and can therefore be consistently identified across individual problems.

\subsection{Recursive Neural Networks} \label{sec:neural_classifier}


A recursive neural network (RvNN) is a network created by \emph{recursively} composing 
a finite set of neural building blocks over a structured input \cite{DBLP:conf/icnn/GollerK96}.
A general neural block is a function $N_\theta : \mathbb{R}^k \to \mathbb{R}^l$  
depending on a vector of parameters $\theta$ that can be optimized during training (see below in Section~\ref{sec:training}). 
  
In our case, the structured input is a clause derivation, i.e.~a DAG
with nodes identified with the derived clauses.
To enable a recursion, an RvNN represents each node $C$ 
by a real vector $v_C$ (of a fixed dimension $n$)
called a (learnable) \emph{embedding}. During training a network learns to embed
the space of derivable clauses into $\mathbb{R}^n$ in some a priori unknown, but still useful way. 

We assume that each initial clause $C$, a leaf of the derivation DAG,
is labeled as belonging to one of the automatically added theory axioms 
or coming from the user input. Let these labels form a finite set of \emph{axiom origin labels} $\mathcal{L}_A$.
Furthermore, let the applicable inference rules that label the internal nodes
of the DAG form a finite set of \emph{inference rule labels} $\mathcal{L}_R$.
The specific building blocks of our neural architecture are the following three (indexed families of) functions: 
\begin{itemize}
\item
	for every axiom label $l \in \mathcal{L}_A$, a nullary \emph{init} function $I_l \in \mathbb{R}^n$ 
	which to an initial clause $C$ labeled by $l$ assigns its embedding $v_C := I_l,$
\item
	for every inference rule $r \in \mathcal{L}_R$, a \emph{deriv} function, $D_r : \mathbb{R}^n \times \cdots \times \mathbb{R}^n \to \mathbb{R}^n$ which
	to a conclusion clause $C_c$ derived by $r$ from premises $(C_1,\ldots,C_k)$ with embeddings $v_{C_1},\ldots,v_{C_k}$ 
	assignes the embedding $v_{C_c} := D_r(v_{C_1},\ldots,v_{C_k})$,
\item
	and, finally, a single \emph{eval} function $E : \mathbb{R}^n \to \mathbb{R}$ 
	which evaluates an embedding $v_C$ such that
	the corresponding clause $C$ is classified as \emph{positive} whenever $E(v_C) \geq t$,
	with the threshold $t$ set, by default, to 0.
\end{itemize}
By recursively composing the init and deriv functions, any derived clause $C$ can be 
assigned an embedding $v_C$ and also evaluated by $E$ 
to see whether the network recommends it as positive, that should be preferred in proof search. 


\subsection{Architecture Details} \label{sec:architecture}


Here we outline the details of our architecture for the benefit of neural network practitioners.
All the used terminology is standard (see, e.g., \cite{DBLP:books/daglib/0040158}).

We realized each init function $I_l$ as an independent learnable vector. Similarly, each deriv function $D_r$
was independently defined. For a rule of arity two, such as resolution, we used:
\begin{equation*} 
D_r(v_1,v_2) = \mathrm{LayerNorm}(y), \, y = W_2^r\cdot x+b^r_2, \, x = \mathrm{ReLU}(W^r_1\cdot[v_1,v_2]+b^r_1), 
\end{equation*}
where $[\cdot,\cdot]$ denotes vector concatenation, $\mathrm{ReLU}$ is the rectified linear unit non-linearity ($f(x)=\max \{0,x\}$) 
applied component-wise, and the learnable matrices $W^r_1,W_2^r$ and vectors $b^r_1,b^r_2$ are such that $x \in \mathbb{R}^{2n}$ and $y \in \mathbb{R}^n$.
(We took inspiration from Sandler et al.~\cite{DBLP:conf/cvpr/SandlerHZZC18} for doubling the embedding size before applying the non-linearity.)
Finally, $\mathrm{LayerNorm}$ is a \emph{layer normalization} \cite{DBLP:journals/corr/BaKH16} module,
without which training often became numerically unstable for deeper derivation DAGs.\footnote{
We also tried to skip $\mathrm{LayerNorm}$ and replace ReLU by the hyperbolic tangent function.
This restores stability, but does not train or classify so well.} 

For unary inference rules, such as factoring, we used an equation analogous to the above, except for the concatenation operation.
We did not need to model an inference rule with a variable number of premises, but one option would be
to arbitrarily ``bracket'' its arguments into a tree of binary applications.


Finally, the eval function was $E(v) = W_2\cdot\mathrm{ReLU}(W_1\cdot v + b)+c$ with trainable 
$W_1 \in \mathbb{R}^{n \times n}, b \in \mathbb{R}^n, W_2 \in \mathbb{R}^{1 \times n},$ and $c\in\mathbb{R}$.

\subsection{Training the Network} \label{sec:training}

To train a network means to find values for the trainable parameters such that 
it accurately classifies the training data and ideally also generalises to unseen future cases.
We follow a standard methodology for training our RvNN. 

In particular, we use the gradient descent (GD) optimization algorithm 
(with the Adam optimiser \cite{DBLP:journals/corr/KingmaB14})
minimising the typical binary cross-entropy \emph{loss}, composed as a sum of contributions, 
for every selected clause $C$, of the form
\[-y_C\cdot\log (\sigma(E(v_C))) -(1-y_C)\cdot\log (1-\sigma(E(v_C))), \]
with $y_C = 1$ for the positive and $y_C = 0$ for the negative examples.


These contributions are weighted such that each derivation DAG 
(corresponding to a prover run on a single problem)
receives equal weight. Moreover, within each DAG 
we re-scale the influence of positive versus the negative 
examples such that these two categories contribute evenly.
The scaling is important as our training data is highly unbalanced (cf. Sect.~\ref{sec:data_prep}).

We split the available successful derivations into a \emph{training} set and a \emph{validation} set,
and only train on the first set using the second to observe generalisation to unseen examples.
As the GD algorithm progresses, iterating over the training data in rounds called \emph{epochs},
we evaluate the loss on the validation set and stop the process early if this loss does not decrease for a specified period.
This \emph{early stopping} criterion was important to produce a model that generalizes well. 

As another form of regularisation, i.e.~a technique for preventing overfitting to the training data,
we employ \emph{dropout} \cite{DBLP:journals/jmlr/SrivastavaHKSS14} (independently for each ``read''
of a clause embedding by one of the deriv or eval functions). Dropout means that at training time
each component $v_i$ 
of the embedding $v$ has a certain probability of being zero-ed out. 
This ``voluntary brain damage'' makes the network more robust as it prevents neurons
from forming too complex co-adaptations \cite{DBLP:journals/jmlr/SrivastavaHKSS14}.

Finally, we experimented with using non-constant learning rates
as suggested by Smith et al.~\cite{DBLP:conf/wacv/Smith17,DBLP:journals/corr/abs-1708-07120}.
In the end, we used a schedule with a linear warmup for the first 50 epochs
followed by a hyperbolic cooldown \cite{DBLP:conf/nips/VaswaniSPUJGKP17} (cf. Fig.~\ref{fig:run40_training} in Sect.~\ref{sec:exper_training}).

%

\subsection{An Abstraction for Compression and Caching} \label{sec:training_pipeline} \label{sec:cache} 

Since our representation of clauses deliberately discards information,
we end up encountering distinct clauses indistinguishable from the perspective of the network.
For example, every initial clause $C$ originating from the input problem 
(as opposed to being added as a theory axiom) receives the same embedding $v_C = I_{\mathit{input}}$.
Indistinguishable clauses also arise as conclusions of an inference 
that can be applied in more than one way to certain premises.

Mathematically, we deal with an equivalence relation $\sim$ on clauses based on ``having the same derivation tree'':
$C_1 \sim C_2 
\leftrightarrow 
\mathit{derivation}(C_1) = \mathit{derivation}(C_2).$
The ``fingerprint'' $\mathit{derivation}(C)$ of a clause could be defined 
as a formal expression recording the derivation history of $C$ using the labels from $\mathcal{L}_A$ as nullary operators
and those from $\mathcal{L}_R$ as operators with arities of the corresponding inference rules.
For example: $\mathit{Resolution}(\mathit{thax\_inverse\_assoc},\mathit{Factoring}(\mathit{input}))$.

We made use of this equivalence in our implementation in two places:
\begin{enumerate}
\item
	When preparing the training data. We ``compressed'' each derivation DAG as a 
	factorisation by $\sim$,
	keeping only one representative of each class. A class containing a positive 
	example was marked as a positive example.


\item
	When interfacing the trained model from the ATP. We cached the embeddings (and evaluated logits)
	for the already encountered clauses under their class identifier.
	Sect.~\ref{sec:lazy_exper} evaluates the effect of this technique.
	
\end{enumerate}




\section{Experiments} \label{sec:experiments}

We implemented the infrastructure for training an RvNN clause derivation classifier (as described in Sect.~\ref{sec:neural_training})
in Python, relying on the PyTorch (version 1.7) library \cite{NEURIPS2019_9015}
and its {Torch\-Script} extension for interfacing the trained model from C++.
We modified the automatic theorem prover \vampire{} (version 4.5.1) to (1) optionally
record to a log file the constructed derivation, including information 
on selected clauses and clauses found in the discovered proof (the \emph{logging-mode}), (2) to be able to load a trained {Torch\-Script} model 
and use it for clause selection guidance under various modes of integration (detailed in Sects.~\ref{sec:integration} and \ref{sec:layered}).\footnote{
Supplementary materials can be found at \url{https://git.io/JtHNl}.}

We took the same subset of \num{20795} problems from the \smtlib{} library \cite{BarFT-SMTLIB} as in previous work \cite{DBLP:conf/cade/Gleiss020}: formed as the largest set of problems 
in a fragment supported by \vampire{},
excluding problems known to be satisfiable
and those provable by \vampire{}'s default strategy
in \SI{10}{\second} either without adding theory axioms or while performing clause selection by age only.

As the baseline strategy $\mathcal{S}$ we took \vampire{}'s implementation of the DISCOUNT saturation loop
under the age-to-weight ratio 1:10 (which typically performs well with DISCOUNT),
keeping all other settings default, including the enabled \avatar{} architecture.
We later enhanced this $\mathcal{S}$ with various forms of guidance. 
All the benchmarking was done using a \SI{10}{\second} time limit.\footnote{
Running on an Intel(R) Xeon(R) Gold 6140 CPUs @ \SI{2.3}{\giga\hertz} server with \SI{500}{\giga\byte} RAM,
using no more than 30 of the available 72 cores to reduce mutual influence.}

\subsection{Data Preparation} \label{sec:data_prep}

During an initial run, the baseline strategy $\mathcal{S}$ was able to solve \num{734} problems
under the \SI{10}{\second} time limit. We collected the corresponding successful derivations 
using the logging-mode (and lifting the time limit, since the logging causes a non-negligible overhead)
and processed them into a form suitable for training a neural model.
The derivations contained approximately \num{5.0} million clauses in total (the overall context),
out of which \num{3.9} million were selected\footnote{
Ancestors of selected clauses are sometimes not selected clauses themselves 
if they arise through immediate simplifications or through reductions.}
(the training examples) and \num{30} thousand of these appeared in a proof (the positive examples).
In these derivations, \vampire{} used 31 distinct theory axioms to facilitate theory reasoning.
Including the ``user input'' label for clauses coming from the actual problem files, 
there were in total 32 distinct labels for the derivation leaves.
In addition, we recorded 15 inference rules, such as resolution, superposition,
backward and forward demodulation or subsumption resolution and including one rule for 
the derivation of a component clause in \avatar{} \cite{DBLP:conf/cav/Voronkov14,DBLP:conf/cade/RegerSV15}.
Thus we obtained 15 distinct labels for the internal nodes.

We compressed these derivations identifying clauses with the same ``abstract derivation history'' dictated by the labels,
as described in Sect.~\ref{sec:training_pipeline}. 
This reduced the derivation set to \num{0.7} million nodes (i.e.~abstracted clauses) in total. 
Out of the \num{734} derivations \num{242} were still larger than \num{1000} nodes (the largest had \num{6426} nodes)
and each of these gave rise to a separate ``mini-batch''. We grouped 
the remaining \num{492} derivations
to obtain an approximate size of \num{1000} nodes per mini-batch (the maximum was 12 original derivations grouped in one mini-batch).
In total, we obtained \num{412} mini-batches and randomly singled out 330 (i.e., \SI{80}{\percent}) of these for training, keeping 82 aside for validation.

\subsection{Training} \label{sec:exper_training}


\begin{figure}[t]

\includegraphics[height=5.3cm,trim={0.4cm 0 0 0},clip]{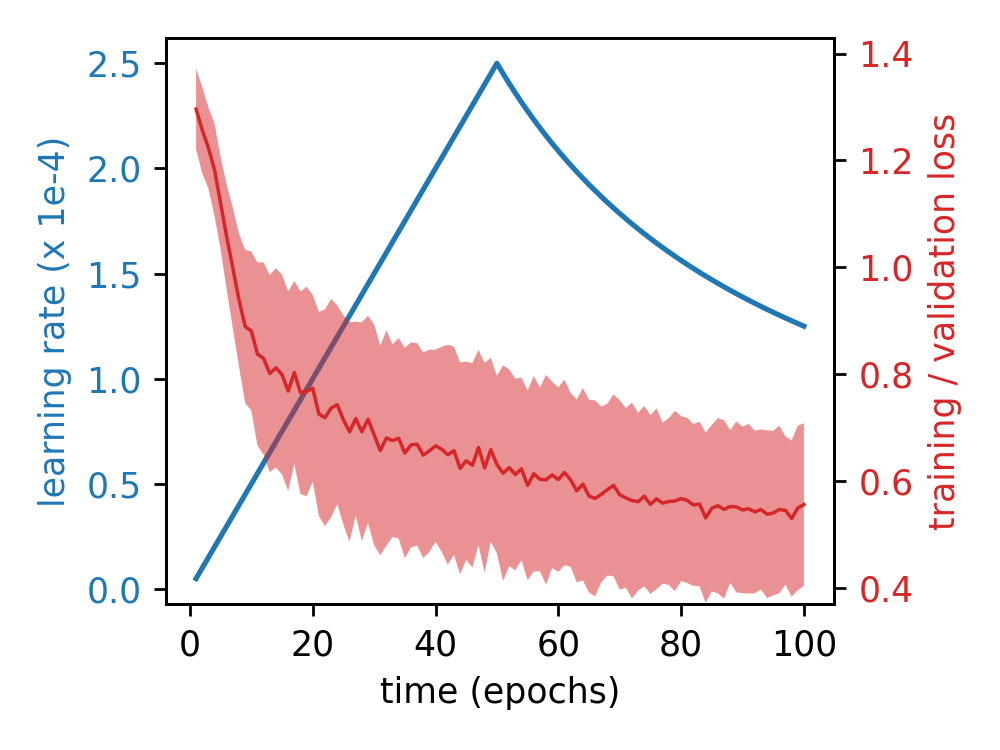}{\hspace*{-0.5em}\includegraphics[height=5.3cm,trim={0.4cm 0 0 0},clip]{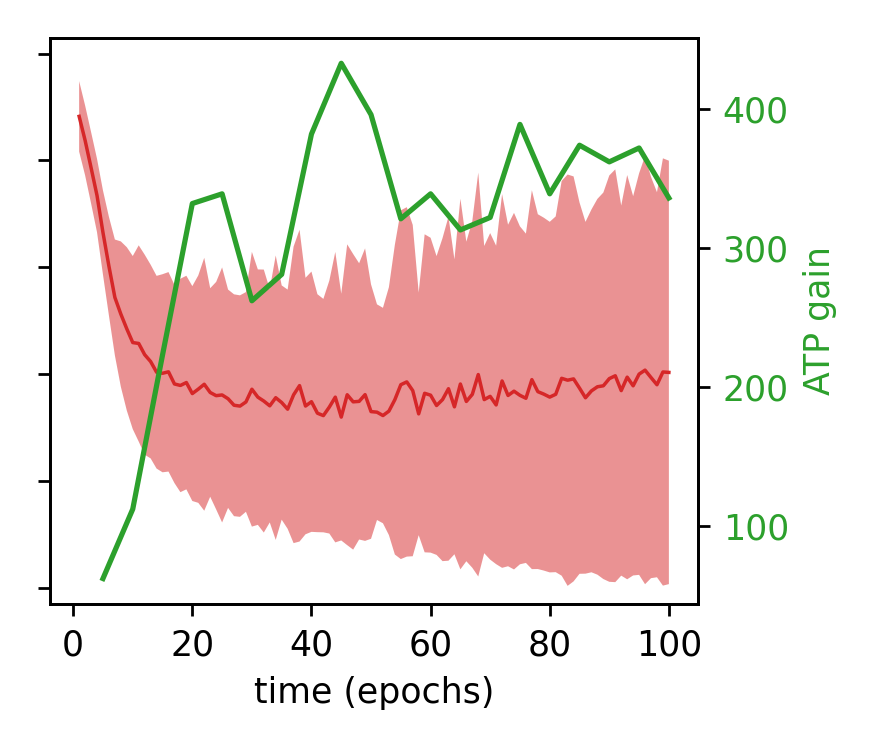}}
\caption{Training the neural model. Red: the training (left) and validation (right) loss as a function training time;
shaded: per problem weighted standard deviations. Blue (left): the supplied non-constant learning rate (cf. Sect.~\ref{sec:training}).
Green (right): in training unseen problems solved by \vampire{} equipped with the corresponding model.}
\label{fig:run40_training}
\end{figure}

Since the size of the training set is relatively small, we instantiated the architecture described in Sect.~\ref{sec:architecture}
with embedding size $n=64$ and dropout probability $p=0.3$. We trained for 100 epochs, with a non-constant learning rate
peaking at $\alpha = \SI{2.5e-4}{}$ in epoch 50. Every epoch we computed the loss on the validation set and 
selected the model which minimizes this quantity. This was the model from epoch 45 in our case,
which we will denote $\mathcal{M}$ here.

The development of the training and validation loss throughout training, as well as that of the learning rate,
is plotted in Fig.~\ref{fig:run40_training}. Additionally, the right side of the figure allows us to 
compare the validation loss---an ML estimate of the model's ability to generalize---with 
the ultimate metric of practical generalization, namely
the number of in-training-unseen problems solved by \vampire{} equipped with the corresponding model for guidance.\footnote{
Integrated using the layered scheme with a second level ratio 2:1 (cf.~Sect.~\ref{sec:integraton_exper}).}
We can see that 
the ``proxy'' (i.e. the minimisation of the validation loss) and the ``target'' (i.e. the maximisation of ATP performance) correspond quite well, 
at least to the degree that we measured the highest ATP gain with the validation-loss-minimizing $\mathcal{M}$.

We remark that this assurance was not cheap to obtain. While the whole 100 epoch training took 45 minutes
to complete (using 20 workers and 1 master process in a parallel training setup), each of the 20 ATP evaluation
data points corresponds to approximately 2 hours of 30 core computation. 




\subsection{Advice Integration} \label{sec:integraton_exper}

In this part of the experiment we tested the various ways of integrating the learnt advice as described in Sects.~\ref{sec:integration} and \ref{sec:layered}.
Let us recall that these are the single queue schemes $\mathcal{M}^{\mathbb{-R}}$ and $\mathcal{M}^{1,0}$
based on the raw logits and the binary decision, respectively, 
their combinations $\mathcal{S}\oplus \mathcal{M}^{\mathbb{-R}}$ and $\mathcal{S}\oplus \mathcal{M}^{1,0}$ 
with the base strategy $\mathcal{S}$ under some second level ratio,
and, finally, $\mathcal{S} \oplus \mathcal{S}[\mathcal{M}^{1}]$, 
the integration of the guidance by the layered clause selection scheme.

\begin{table}[t]
\caption{Performance results of various forms of integrating the model advice.}
\label{tab:integraton}
\centering
\begin{tabular}{cc|c|cc|cc}
strategy & ratio & $\mathcal{M}$ eval. time\% & \#solved & (percent $\mathcal{S}$) & gained & lost  \\
\hline
$\mathcal{S}$ & $-$ & \phantom{0}\SI{0}{\percent} & \phantom{0}756 & \SI{100}{\percent} & \phantom{0}26 & \phantom{00}4 \\
\hline
$\mathcal{M}^{\mathbb{-R}}$ & $-$ & \SI{25}{\percent} & \phantom{00}55 & \phantom{00}\SI{7}{\percent} & \phantom{0}25 & 704  \\
$\mathcal{M}^{1,0}$ & $-$ & \SI{13}{\percent} & {\bf \phantom{0}190} & \phantom{0}\SI{25}{\percent} & {\bf \phantom{0}30} & {\bf 574} \\
\hline
$\mathcal{S}\oplus \mathcal{M}^{\mathbb{-R}}$ & \phantom{0}5:1\phantom{0} & \SI{57}{\percent} & {\bf \phantom{0}543} & \phantom{0}\SI{71}{\percent} & {\bf \phantom{0}86} & {\bf 277} \\
 & \phantom{0}2:1\phantom{0} & \SI{48}{\percent} & \phantom{0}445 & \phantom{0}\SI{58}{\percent} & \phantom{0}78 & 367 \\
 & \phantom{0}1:1\phantom{0} & \SI{41}{\percent} & \phantom{0}335 & \phantom{0}\SI{44}{\percent} & \phantom{0}54 & 453 \\
 & \phantom{0}1:2\phantom{0} & \SI{32}{\percent} & \phantom{0}248 & \phantom{0}\SI{32}{\percent} & \phantom{0}39 & 525 \\
 & \phantom{0}1:5\phantom{0} & \SI{32}{\percent} & \phantom{0}140 & \phantom{0}\SI{18}{\percent} & \phantom{0}28 & 622 \\
\hline
$\mathcal{S}\oplus \mathcal{M}^{1,0}$ & 10:1\phantom{0} & \SI{11}{\percent} & {\bf \phantom{0}686} & \phantom{0}\SI{90}{\percent} & \phantom{0}80 & {\bf 128}\\
 & \phantom{0}2:1\phantom{0} & \SI{14}{\percent} & \phantom{0}602 & \phantom{0}\SI{79}{\percent} & 112 & 244 \\
 & \phantom{0}1:1\phantom{0} & \SI{14}{\percent} & \phantom{0}555 & \phantom{0}\SI{73}{\percent} & 111 & 290 \\
 & \phantom{0}1:2\phantom{0} & \SI{14}{\percent} & \phantom{0}519 & \phantom{0}\SI{68}{\percent} & {\bf 132} & 347 \\
 & \phantom{0}1:10 & \SI{14}{\percent} & \phantom{0}520 & \phantom{0}\SI{68}{\percent} & {\bf 132} & 346 \\
\hline
$\mathcal{S}\oplus \mathcal{S}[\mathcal{M}^{1}]$ & \phantom{0}2:1\phantom{0} & \SI{27}{\percent} & \phantom{0}855 & \SI{113}{\percent} & 210 & {\bf \phantom{0}89} \\
 & \phantom{0}1:1\phantom{0} & \SI{32}{\percent} & 1032 & \SI{136}{\percent} & 411 &  113 \\
 & \phantom{0}1:2\phantom{0} & \SI{33}{\percent} & {\bf 1036}  & \SI{137}{\percent} & {\bf 430} & 128 \\
 & \phantom{0}1:3\phantom{0} & \SI{30}{\percent} & 1026 & \SI{135}{\percent} & 428 & 136 \\ 
 & \phantom{0}1:5\phantom{0} & \SI{25}{\percent} & \phantom{0}989 & \SI{130}{\percent} & 405 & 150 \\
\end{tabular}
\end{table}


Our results are shown in Table~\ref{tab:integraton}.
It starts by reporting on the performance of the baseline strategy $\mathcal{S}$
and then compares it to the other strategies (the gained and lost columns are w.r.t. the original run of $\mathcal{S}$).\footnote{
We had to switch to a different machine after producing the training data.
There, a rerun of $\mathcal{S}$ gave a slightly better performance than the 
\num{734} solved problems 
used for training.
We still used the original run's results to compute the gained and lost values here;
the percentage solved is with respect to the new run of $\mathcal{S}$.}
We can see that the two single queue approaches are quite weak,
with the better $\mathcal{M}^{1,0}$ solving only \SI{25}{\percent} of the baseline.
Nor can the combination $\mathcal{S}\oplus \mathcal{M}^{\mathbb{-R}}$ be considered a success,
as it only solves more problems when less and less advice is taken,
seemingly approaching the performance of $\mathcal{S}$ from below.
This trend repeats with $\mathcal{S}\oplus \mathcal{M}^{1,0}$, 
although here an interesting number of problems not solved by the baseline 
is gained by strategies which rely on the advice more than half of the time.

With our model $\mathcal{M}$, only the layered clause selection integration $\mathcal{S}\oplus \mathcal{S}[\mathcal{M}^{1}]$
is able to improve on the performance of the baseline strategy $\mathcal{S}$.
In fact, it improves on it very significantly: with the second level ratio of 1:2 we achieve \SI{137}{\percent} performance of the baseline
and gain 430 problems unsolved by $\mathcal{S}$. 

\subsection{Evaluation Speed, Lazy Evaluation, and Abstraction Caching} \label{sec:lazy_exper}

Table~\ref{tab:integraton} also shows the percentage of computation time 
the individual strategies spent evaluating the advice, i.e.~interfacing $\mathcal{M}$. 

A word of warning first. These number are hard to interpret across different strategies.
It is because different guidance steers the prover to different parts of the search space.
For example, notice the seemingly paradoxical situation most pronounced with $\mathcal{S}\oplus \mathcal{M}^{\mathbb{-R}}$,
where the more often is the advice from $\mathcal{M}$ nominally requested, the less time the prover spends interfacing $\mathcal{M}$. 
Looking closely at a few problems, we discovered that in strategies relying a lot on $\mathcal{M}^{\mathbb{-R}}$,
such as $\mathcal{S}\oplus \mathcal{M}^{\mathbb{-R}}$ under the ratio 1:5, most of the time is spent performing forward subsumption.
An explanation is that the guidance becomes increasingly bad and the prover slows down, processing larger and larger clauses
for which the subsumption checks are expensive and dominate the runtime.\footnote{A similar experience with bad guidance has been made by the authors of ENIGMA.} 

\begin{table}[t]
\caption{Performance decrease caused by turning off abstraction caching and lazy evaluation, and both;
demonstrated on $\mathcal{S}\oplus \mathcal{S}[\mathcal{M}^{1}]$ under the second level ratio 1:2.}
\label{tab:lazy_abst_perf}
\centering
\begin{tabular}{l|c|cc}
 & $\mathcal{M}$ eval. time\% & \#solved & (percent $\mathcal{S}$) \\
\hline
both techniques enabled & \SI{33}{\percent} & 1036  & \SI{137}{\percent}\\
without abstraction caching & \SI{45}{\percent} & 1007 & \SI{133}{\percent}  \\  
without lazy evaluation & \SI{58}{\percent} & \phantom{0}905 & \SI{119}{\percent}  \\  
both techniques disabled & \SI{73}{\percent} & \phantom{0}782 & \SI{103}{\percent}  \\  
\end{tabular}
\end{table}

When the guidance is the same, however, we can use the eval.~time percentage to estimate the efficiency of the integration.
The results shown in Table~\ref{tab:integraton} were obtained using both lazy evaluation\footnote{
With the exception of the $\mathcal{M}^{\mathbb{-R}}$ guidance, with which it is incompatible.}
and abstraction caching (as described in sections \ref{sec:lazy} and \ref{sec:cache}). 
Taking the best performing $\mathcal{S}\oplus \mathcal{S}[\mathcal{M}^{1}]$ under the second level ratio 1:2,
we selectively disabled: first abstraction caching, then lazy evaluation and finally both techniques,
obtaining the values shown in Table~\ref{tab:lazy_abst_perf}.

We can see that the techniques considerably contribute to the overall performance.
Indeed, without them \vampire{} would spend the whole \SI{73}{\percent} of computation time evaluating the network
(compared to only \SI{33}{\percent}) and the strategy would barely match (with \SI{103}{\percent}) the performance of the baseline $\mathcal{S}$.



\subsection{Positive Bias}


Two important characteristics, from a machine learning perspective, of an obtained model 
are the \emph{true positive rate} (TPR) (also called sensitivity) and the \emph{true negative rate} (TNR) (also specificity).
TPR is defined as the fraction of positively labeled examples which the model also classifies as such.
TNR is, analogously, the fraction of negatively labeled examples.
Our model $\mathcal{M}$ achieves (on the validation set) \SI{86}{\percent} TPR and \SI{81}{\percent} TNR.

\begin{figure}[t]
\includegraphics[height=5.3cm]{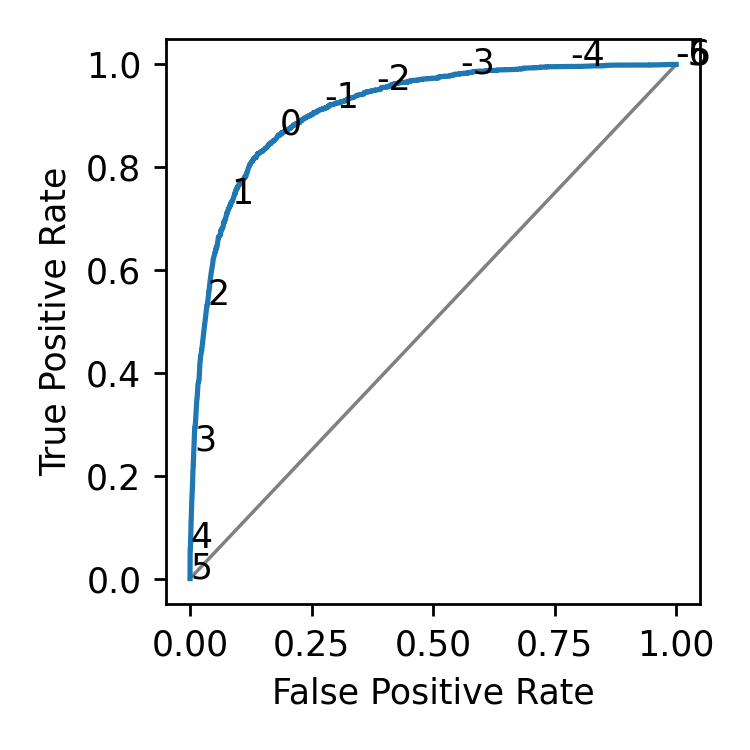}{\hspace*{-0.5em}\includegraphics[height=5.3cm]{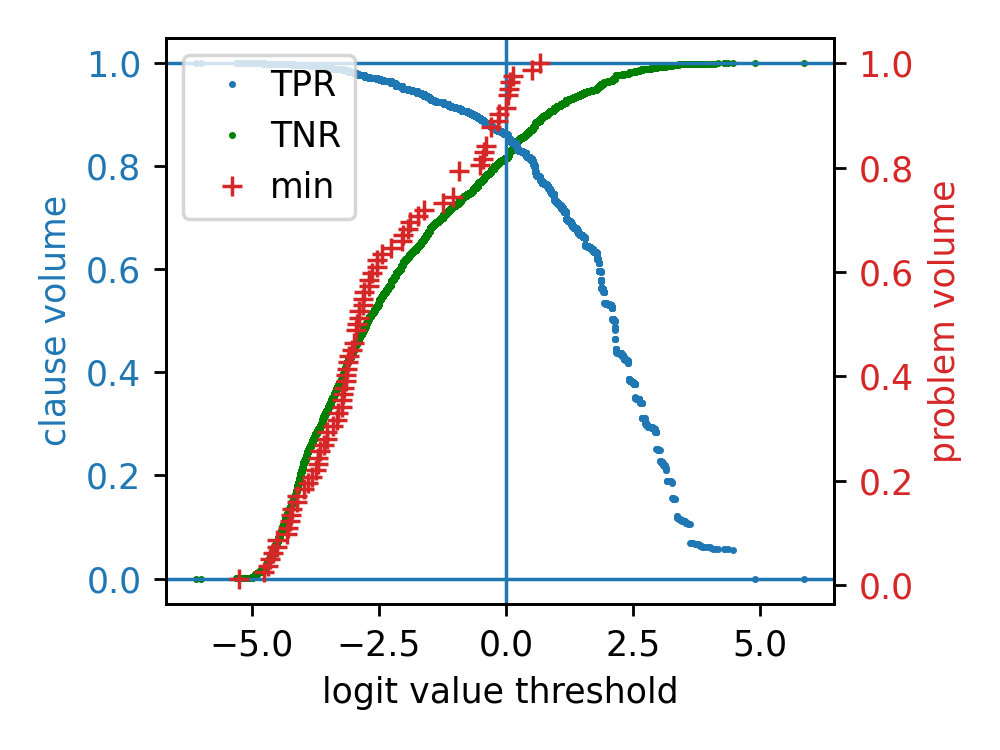}}
\caption{The receiver operating characteristic curve (left) and a related plot with explicit threshold (right) 
for the selected model $\mathcal{M}$; 
both based on validation data.}
\label{fig:ml_rates}
\end{figure}

The final judgement of a neural classifier follows from a comparison to a threshold value $t$, 
set by default to $t = 0$ (recall Sect.~\ref{sec:neural_classifier}). Changing this threshold
allows us to trade TPR for TNR and vice versa in straightforward way. The interdependence 
of these two values on the varied threshold is  traditionally captured 
by the so called \emph{receiver operating characteristic} (ROC) curve,
shown for our model in Fig.~\ref{fig:ml_rates} (left). The tradition dictates that 
the $x$ axis be labeled by the false positive rate (FPR) (also called fall-out) which is simply $1-\mathrm{TNR}$.
Under such presentation, one generally strives to pick a threshold value at which the curve is the closest 
to the upper left corner of the plot.\footnote{Minimizing the standard cross entropy loss 
should actually automatically ``bring the curve'' close to that corner for the threshold $t=0$.}
However, this is not necessarily the best configuration for every application.

In the Fig.~\ref{fig:ml_rates} (right), we ``decompose'' the ROC curve 
by using the threshold $t$ for the independent axis $x$.
We also highlight, for every problem (again, in the validation set),
what is the minimal logit value across all positively labeled examples
belonging to that problem. In other words, what is the logit of the ``least positively classified''
clause from the problem's proof. We can see that for the majority of the problems
these minima are below the threshold $t=0$. This means that for those problems at least one clause from the original proof 
is getting classified as negative by $\mathcal{M}$ under $t=0$.


\begin{table}[t]
\caption{The performance of $\mathcal{S}\oplus \mathcal{S}[\mathcal{M}^{1}]$ under the second level ratio 1:2
while changing the logit threshold. A smaller threshold means more clauses classified as positive.  }
\label{tab:positive_bias}
\centering
\begin{tabular}{c|cc|cc}
threshold & \#solved & (percent $\mathcal{S}$) & gained & lost  \\
$-0.50$ & 1063 & \SI{140}{\percent} & 427 & \phantom{0}{\bf 98}  \\
$-0.25$ & {\bf 1066} & \SI{141}{\percent} & {\bf 439} & 107  \\
$\phantom{-}0.00$ & 1036 & \SI{137}{\percent} &  430 & 128 \\
$\phantom{-}0.25$ & \phantom{0}945 &  \SI{125}{\percent} & 375 & 164  \\
$\phantom{-}0.50$ & \phantom{0}825 &  \SI{109}{\percent} & 278 & 187  \\
\end{tabular}
\end{table}

These observations motivated us to experiment with non-zero values of the threshold in an ATP evaluation.
Particularly promising seemed the use of a threshold $t$ smaller than zero
with the intention of classifying more clauses as positive. 
The results of the experiment are in shown Table~\ref{tab:positive_bias}.
Indeed, we could further improve the best performing
strategy from Table~\ref{tab:integraton} with both $t=-0.25$ and $t=-0.5$. It can be seen that smaller
values lead to fewer problems lost, but even the ATP gain is better with $t=-0.25$ than with the default $t=0$,
leading to the overall best improvement of \SI{141}{\percent} with respect to the baseline $\mathcal{S}$.

\subsection{Learning from Guided Proofs and Negative Mining}

As previously unsolved problems get proven with the help of the trained guidance,
the new proofs can 
be used to enrich the training set and potentially help obtaining
even better models. This idea of alternating the training and the ATP evaluation
steps in a reinforcing \emph{loop} has been proposed and successfully realized by the authors of 
ENIGMA on the Mizar dataset \cite{DBLP:conf/itp/JakubuvU19}. Here we propose an enhancement 
of the idea and repeat an analogous experiment in our setting.

By collecting proofs discovered by a selection of 8 different configurations 
tested in the previous sections, we grew our set of solved problems from
\num{734} to \num{1528}. We decided to keep one proof per problem, strictly
extending the original training set. We then repeated the same training procedure
as described in Sect.~\ref{sec:exper_training} on this new set and 
on an extension of this set obtained as follows.

\paragraph{Negative mining:}
We suspected that the successful derivations obtained
with the help of $\mathcal{M}$ might not contain enough ``typical wrong decisions''
from the perspective of $\mathcal{S}$ to provide for good enough training. 
We therefore logged the \emph{failing} runs of $\mathcal{S}$ on
the $(1528-734)$ problems only solved by one of the guided strategies
and augmented the corresponding derivations with these.\footnote{
Negative mining has, for instance, been previously used when training
deep models for the premise selection task \cite{DBLP:conf/nips/IrvingSAECU16}.
}

\begin{table}[t]
\caption{The performance of new models learned from guided proofs.
    $\mathcal{U}$ is the set of \num{1528} problems used for the training. 
    The gained and lost counts are here w.r.t. $\mathcal{U}$.}
\label{tab:looping}
\centering
\begin{tabular}{r|ccc|cc}
 & \#solved & (percent $\mathcal{S}$) & (percent $|\mathcal{U}|$) & gained & lost  \\
plain                & 1268 & \SI{167}{\percent} & \SI{82}{\percent} & \phantom{0}90 & 350 \\
with negative mining & {\bf 1394} & {\bf \SI{184}{\percent}} & \SI{91}{\percent} & {\bf 140} & {\bf 274} \\
\end{tabular}
\end{table}

Table~\ref{tab:looping} confirms\footnote{The ATP eval was again 
integrating via $\mathcal{S}\oplus \mathcal{S}[\mathcal{M}^{1}]$ under the second level ratio 1:2.} 
that negative mining indeed helps to produce a better model.
Mainly, however, it shows that training from additional derivations 
further dramatically improves the performance of the obtained strategy.






\section{Conclusion} \label{sec:conclusion}

We revisited the topic of ENIGMA-style clause selection guidance by a machine learned binary classifier and
proposed four improvements to previous work: (1) the use of layered clause selection for integrating the advice,
(2) the lazy evaluation trick to reduce the overhead of interfacing a potentially expensive model,
(3) the ``positive bias'' idea suggesting to be really careful not to discard potentially useful clauses,
and (4) the ``negative mining'' technique to provide enough negative examples when learning
from proofs obtained with previous guidance.

We have also shown that a strong advice can be obtained by looking just at the derivation history to discriminate a clause.
The automatically discovered neural guidance significantly improves upon the human-engineered heuristic 
\cite{DBLP:conf/cade/Gleiss020} under identical conditions. Rerunning $\mathcal{S}$ with the theory 
heuristic enabled in its default form \cite{DBLP:conf/ijcai/Gleiss020} resulted here in \num{816} (\SI{107}{\percent}) solved problems.


By deliberately focusing of the representation of clauses by their derivations, we obtained 
some nice properties, such as relative speed of evaluation. 
%
However, in situations where theory reasoning by automatically added
theory axioms is not prevalent, such as on most of the TPTP library,
we expect guidance based on derivations with just a single axiom origin label, the $\mathit{input}$, to be quite weak. 

Still, we see a great opportunity in using statistical methods for analyzing 
ATP behaviour; not only for improving prover performance with a black box guidance,
but also as a tool for discovering regularities that could be exploited 
to improve our understanding of the technology on a deeper level. 





\section*{Acknowledgement}
This work was supported by the Czech Science Foundation project 20-06390Y 
and the project RICAIP no. 857306 under the EU-H2020 programme. We also thank
the anonymous reviewers for suggesting numerous improvements.

%
%
\bibliographystyle{splncs04}
\bibliography{main}

\begin{thebibliography}{10}
\providecommand{\url}[1]{\texttt{#1}}
\providecommand{\urlprefix}{URL }
\providecommand{\doi}[1]{https://doi.org/#1}

\bibitem{DBLP:journals/corr/abs-2006-11259}
Ayg{\"{u}}n, E., Ahmed, Z., Anand, A., Firoiu, V., Glorot, X., Orseau, L.,
  Precup, D., Mourad, S.: Learning to prove from synthetic theorems. CoRR
  \textbf{abs/2006.11259} (2020), \url{https://arxiv.org/abs/2006.11259}

\bibitem{DBLP:journals/corr/BaKH16}
Ba, L.J., Kiros, J.R., Hinton, G.E.: Layer normalization. CoRR
  \textbf{abs/1607.06450} (2016), \url{http://arxiv.org/abs/1607.06450}

\bibitem{DBLP:journals/logcom/BachmairG94}
Bachmair, L., Ganzinger, H.: Rewrite-based equational theorem proving with
  selection and simplification. J. Log. Comput.  \textbf{4}(3),  217--247
  (1994). \doi{10.1093/logcom/4.3.217},
  \url{https://doi.org/10.1093/logcom/4.3.217}

\bibitem{DBLP:books/el/RV01/BachmairG01}
Bachmair, L., Ganzinger, H.: Resolution theorem proving. In: Robinson and
  Voronkov  \cite{DBLP:books/el/RobinsonV01}, pp. 19--99.
  \doi{10.1016/b978-044450813-3/50004-7},
  \url{https://doi.org/10.1016/b978-044450813-3/50004-7}

\bibitem{BarFT-SMTLIB}
Barrett, C., Fontaine, P., Tinelli, C.: {The Satisfiability Modulo Theories
  Library (SMT-LIB)}. {\tt www.SMT-LIB.org} (2016)

\bibitem{DBLP:conf/vmcai/BradleyMS06}
Bradley, A.R., Manna, Z., Sipma, H.B.: What's decidable about arrays? In:
  Emerson, E.A., Namjoshi, K.S. (eds.) Verification, Model Checking, and
  Abstract Interpretation, 7th International Conference, {VMCAI} 2006,
  Charleston, SC, USA, January 8-10, 2006, Proceedings. Lecture Notes in
  Computer Science, vol.~3855, pp. 427--442. Springer (2006).
  \doi{10.1007/11609773\_28}, \url{https://doi.org/10.1007/11609773\_28}

\bibitem{DBLP:conf/cade/ChvalovskyJ0U19}
Chvalovsk{\'{y}}, K., Jakubuv, J., Suda, M., Urban, J.: {ENIGMA-NG:} efficient
  neural and gradient-boosted inference guidance for {E}. In: Fontaine
  \cite{DBLP:conf/cade/2019}, pp. 197--215.
  \doi{10.1007/978-3-030-29436-6\_12},
  \url{https://doi.org/10.1007/978-3-030-29436-6\_12}

\bibitem{DS1996b}
Denzinger, J., Schulz, S.: {Learning Domain Knowledge to Improve Theorem
  Proving}. In: McRobbie, M., Slaney, J. (eds.) Proc.\ of the 13th CADE, New
  Brunswick. pp. 62--76. No.~1104 in LNAI, Springer (1996)

\bibitem{DBLP:conf/cade/2019}
Fontaine, P. (ed.): Automated Deduction - {CADE} 27 - 27th International
  Conference on Automated Deduction, Natal, Brazil, August 27-30, 2019,
  Proceedings, Lecture Notes in Computer Science, vol. 11716. Springer (2019).
  \doi{10.1007/978-3-030-29436-6},
  \url{https://doi.org/10.1007/978-3-030-29436-6}

\bibitem{DBLP:conf/ijcai/Gleiss020}
Gleiss, B., Suda, M.: Layered clause selection for saturation-based theorem
  proving. In: Fontaine, P., Korovin, K., Kotsireas, I.S., R{\"{u}}mmer, P.,
  Tourret, S. (eds.) Joint Proceedings of the 7th Workshop on Practical Aspects
  of Automated Reasoning {(PAAR)} and the 5th Satisfiability Checking and
  Symbolic Computation Workshop (SC-Square) Workshop, 2020 co-located with the
  10th International Joint Conference on Automated Reasoning {(IJCAR} 2020),
  Paris, France, June-July, 2020 (Virtual). {CEUR} Workshop Proceedings,
  vol.~2752, pp. 34--52. CEUR-WS.org (2020),
  \url{http://ceur-ws.org/Vol-2752/paper3.pdf}

\bibitem{DBLP:conf/cade/Gleiss020}
Gleiss, B., Suda, M.: Layered clause selection for theory reasoning - (short
  paper). In: Peltier, N., Sofronie{-}Stokkermans, V. (eds.) Automated
  Reasoning - 10th International Joint Conference, {IJCAR} 2020, Paris, France,
  July 1-4, 2020, Proceedings, Part {I}. Lecture Notes in Computer Science,
  vol. 12166, pp. 402--409. Springer (2020).
  \doi{10.1007/978-3-030-51074-9\_23},
  \url{https://doi.org/10.1007/978-3-030-51074-9\_23}

\bibitem{DBLP:conf/icnn/GollerK96}
Goller, C., K{\"{u}}chler, A.: Learning task-dependent distributed
  representations by backpropagation through structure. In: Proceedings of
  International Conference on Neural Networks (ICNN'96), Washington, DC, USA,
  June 3-6, 1996. pp. 347--352. {IEEE} (1996). \doi{10.1109/ICNN.1996.548916},
  \url{https://doi.org/10.1109/ICNN.1996.548916}

\bibitem{DBLP:books/daglib/0040158}
Goodfellow, I.J., Bengio, Y., Courville, A.C.: Deep Learning. Adaptive
  computation and machine learning, {MIT} Press (2016),
  \url{http://www.deeplearningbook.org/}

\bibitem{DBLP:conf/nips/IrvingSAECU16}
Irving, G., Szegedy, C., Alemi, A.A., E{\'{e}}n, N., Chollet, F., Urban, J.:
  Deepmath - deep sequence models for premise selection. In: Lee, D.D.,
  Sugiyama, M., von Luxburg, U., Guyon, I., Garnett, R. (eds.) Advances in
  Neural Information Processing Systems 29: Annual Conference on Neural
  Information Processing Systems 2016, December 5-10, 2016, Barcelona, Spain.
  pp. 2235--2243 (2016),
  \url{https://proceedings.neurips.cc/paper/2016/hash/f197002b9a0853eca5e046d9ca4663d5-Abstract.html}

\bibitem{DBLP:conf/cade/JakubuvCOP0U20}
Jakubuv, J., Chvalovsk{\'{y}}, K., Ols{\'{a}}k, M., Piotrowski, B., Suda, M.,
  Urban, J.: {ENIGMA} anonymous: Symbol-independent inference guiding machine
  (system description). In: Peltier, N., Sofronie{-}Stokkermans, V. (eds.)
  Automated Reasoning - 10th International Joint Conference, {IJCAR} 2020,
  Paris, France, July 1-4, 2020, Proceedings, Part {II}. Lecture Notes in
  Computer Science, vol. 12167, pp. 448--463. Springer (2020).
  \doi{10.1007/978-3-030-51054-1\_29},
  \url{https://doi.org/10.1007/978-3-030-51054-1\_29}

\bibitem{DBLP:conf/mkm/JakubuvU17}
Jakubuv, J., Urban, J.: {ENIGMA:} efficient learning-based inference guiding
  machine. In: Geuvers, H., England, M., Hasan, O., Rabe, F., Teschke, O.
  (eds.) Intelligent Computer Mathematics - 10th International Conference,
  {CICM} 2017, Edinburgh, UK, July 17-21, 2017, Proceedings. Lecture Notes in
  Computer Science, vol. 10383, pp. 292--302. Springer (2017).
  \doi{10.1007/978-3-319-62075-6\_20},
  \url{https://doi.org/10.1007/978-3-319-62075-6\_20}

\bibitem{DBLP:conf/mkm/JakubuvU18}
Jakubuv, J., Urban, J.: Enhancing {ENIGMA} given clause guidance. In: Rabe, F.,
  Farmer, W.M., Passmore, G.O., Youssef, A. (eds.) Intelligent Computer
  Mathematics - 11th International Conference, {CICM} 2018, Hagenberg, Austria,
  August 13-17, 2018, Proceedings. Lecture Notes in Computer Science, vol.
  11006, pp. 118--124. Springer (2018). \doi{10.1007/978-3-319-96812-4\_11},
  \url{https://doi.org/10.1007/978-3-319-96812-4\_11}

\bibitem{DBLP:conf/itp/JakubuvU19}
Jakubuv, J., Urban, J.: Hammering mizar by learning clause guidance (short
  paper). In: Harrison, J., O'Leary, J., Tolmach, A. (eds.) 10th International
  Conference on Interactive Theorem Proving, {ITP} 2019, September 9-12, 2019,
  Portland, OR, {USA}. LIPIcs, vol.~141, pp. 34:1--34:8. Schloss Dagstuhl -
  Leibniz-Zentrum f{\"{u}}r Informatik (2019).
  \doi{10.4230/LIPIcs.ITP.2019.34},
  \url{https://doi.org/10.4230/LIPIcs.ITP.2019.34}

\bibitem{DBLP:journals/corr/KingmaB14}
Kingma, D.P., Ba, J.: Adam: {A} method for stochastic optimization. In: Bengio,
  Y., LeCun, Y. (eds.) 3rd International Conference on Learning
  Representations, {ICLR} 2015, San Diego, CA, USA, May 7-9, 2015, Conference
  Track Proceedings (2015), \url{http://arxiv.org/abs/1412.6980}

\bibitem{DBLP:conf/popl/KovacsRV17}
Kov{\'{a}}cs, L., Robillard, S., Voronkov, A.: Coming to terms with quantified
  reasoning. In: Castagna, G., Gordon, A.D. (eds.) Proceedings of the 44th
  {ACM} {SIGPLAN} Symposium on Principles of Programming Languages, {POPL}
  2017, Paris, France, January 18-20, 2017. pp. 260--270. {ACM} (2017).
  \doi{10.1145/3009837}, \url{http://dl.acm.org/citation.cfm?id=3009887}

\bibitem{DBLP:conf/cav/KovacsV13}
Kov{\'{a}}cs, L., Voronkov, A.: First-order theorem proving and {Vampire}. In:
  Sharygina, N., Veith, H. (eds.) Computer Aided Verification - 25th
  International Conference, {CAV} 2013, Saint Petersburg, Russia, July 13-19,
  2013. Proceedings. Lecture Notes in Computer Science, vol.~8044, pp. 1--35.
  Springer (2013). \doi{10.1007/978-3-642-39799-8\_1},
  \url{https://doi.org/10.1007/978-3-642-39799-8\_1}

\bibitem{DBLP:conf/lpar/LoosISK17}
Loos, S.M., Irving, G., Szegedy, C., Kaliszyk, C.: Deep network guided proof
  search. In: Eiter, T., Sands, D. (eds.) LPAR-21, 21st International
  Conference on Logic for Programming, Artificial Intelligence and Reasoning,
  Maun, Botswana, May 7-12, 2017. EPiC Series in Computing, vol.~46, pp.
  85--105. EasyChair (2017),
  \url{https://easychair.org/publications/paper/ND13}

\bibitem{DBLP:books/el/RV01/NieuwenhuisR01}
Nieuwenhuis, R., Rubio, A.: Paramodulation-based theorem proving. In: Robinson
  and Voronkov  \cite{DBLP:books/el/RobinsonV01}, pp. 371--443.
  \doi{10.1016/b978-044450813-3/50009-6},
  \url{https://doi.org/10.1016/b978-044450813-3/50009-6}

\bibitem{DBLP:conf/ecai/OlsakKU20}
Ols{\'{a}}k, M., Kaliszyk, C., Urban, J.: Property invariant embedding for
  automated reasoning. In: Giacomo, G.D., Catal{\'{a}}, A., Dilkina, B.,
  Milano, M., Barro, S., Bugar{\'{\i}}n, A., Lang, J. (eds.) {ECAI} 2020 - 24th
  European Conference on Artificial Intelligence, 29 August-8 September 2020,
  Santiago de Compostela, Spain, August 29 - September 8, 2020 - Including 10th
  Conference on Prestigious Applications of Artificial Intelligence {(PAIS}
  2020). Frontiers in Artificial Intelligence and Applications, vol.~325, pp.
  1395--1402. {IOS} Press (2020). \doi{10.3233/FAIA200244},
  \url{https://doi.org/10.3233/FAIA200244}

\bibitem{NEURIPS2019_9015}
Paszke, A., Gross, S., Massa, F., Lerer, A., Bradbury, J., Chanan, G., Killeen,
  T., Lin, Z., Gimelshein, N., Antiga, L., Desmaison, A., Kopf, A., Yang, E.,
  DeVito, Z., Raison, M., Tejani, A., Chilamkurthy, S., Steiner, B., Fang, L.,
  Bai, J., Chintala, S.: Pytorch: An imperative style, high-performance deep
  learning library. In: Wallach, H., Larochelle, H., Beygelzimer, A.,
  d\textquotesingle Alch\'{e}-Buc, F., Fox, E., Garnett, R. (eds.) Advances in
  Neural Information Processing Systems 32, pp. 8024--8035. Curran Associates,
  Inc. (2019),
  \url{http://papers.neurips.cc/paper/9015-pytorch-an-imperative-style-high-performance-deep-learning-library.pdf}

\bibitem{DBLP:conf/cade/RegerSV15}
Reger, G., Suda, M., Voronkov, A.: Playing with {AVATAR}. In: Felty, A.P.,
  Middeldorp, A. (eds.) Automated Deduction - {CADE-25} - 25th International
  Conference on Automated Deduction, Berlin, Germany, August 1-7, 2015,
  Proceedings. Lecture Notes in Computer Science, vol.~9195, pp. 399--415.
  Springer (2015). \doi{10.1007/978-3-319-21401-6\_28},
  \url{https://doi.org/10.1007/978-3-319-21401-6\_28}

\bibitem{DBLP:journals/jsc/RiazanovV03}
Riazanov, A., Voronkov, A.: Limited resource strategy in resolution theorem
  proving. J. Symb. Comput.  \textbf{36}(1-2),  101--115 (2003).
  \doi{10.1016/S0747-7171(03)00040-3},
  \url{https://doi.org/10.1016/S0747-7171(03)00040-3}

\bibitem{DBLP:books/el/RobinsonV01}
Robinson, J.A., Voronkov, A. (eds.): Handbook of Automated Reasoning (in 2
  volumes). Elsevier and {MIT} Press (2001),
  \url{https://www.sciencedirect.com/book/9780444508133/handbook-of-automated-reasoning}

\bibitem{DBLP:conf/cvpr/SandlerHZZC18}
Sandler, M., Howard, A.G., Zhu, M., Zhmoginov, A., Chen, L.: Mobilenetv2:
  Inverted residuals and linear bottlenecks. In: 2018 {IEEE} Conference on
  Computer Vision and Pattern Recognition, {CVPR} 2018, Salt Lake City, UT,
  USA, June 18-22, 2018. pp. 4510--4520. {IEEE} Computer Society (2018).
  \doi{10.1109/CVPR.2018.00474},
  \url{http://openaccess.thecvf.com/content\_cvpr\_2018/html/Sandler\_MobileNetV2\_Inverted\_Residuals\_CVPR\_2018\_paper.html}

\bibitem{Schulz:Diss-2000}
Schulz, S.: {Learning Search Control Knowledge for Equational Deduction}.
  No.~230 in DISKI, Akademische Verlagsgesellschaft Aka GmbH Berlin (2000)

\bibitem{SCV:CADE-2019}
Schulz, S., Cruanes, S., Vukmirovi{\'c}, P.: Faster, higher, stronger: {E} 2.3.
  In: Fontaine, P. (ed.) Proc.\ of the 27th CADE, Natal, Brasil. pp. 495--507.
  No. 11716 in LNAI, Springer (2019)

\bibitem{DBLP:conf/cade/SchulzM16}
Schulz, S., M{\"{o}}hrmann, M.: Performance of clause selection heuristics for
  saturation-based theorem proving. In: Olivetti, N., Tiwari, A. (eds.)
  Automated Reasoning - 8th International Joint Conference, {IJCAR} 2016,
  Coimbra, Portugal, June 27 - July 2, 2016, Proceedings. Lecture Notes in
  Computer Science, vol.~9706, pp. 330--345. Springer (2016).
  \doi{10.1007/978-3-319-40229-1\_23},
  \url{https://doi.org/10.1007/978-3-319-40229-1\_23}

\bibitem{DBLP:conf/wacv/Smith17}
Smith, L.N.: Cyclical learning rates for training neural networks. In: 2017
  {IEEE} Winter Conference on Applications of Computer Vision, {WACV} 2017,
  Santa Rosa, CA, USA, March 24-31, 2017. pp. 464--472. {IEEE} Computer Society
  (2017). \doi{10.1109/WACV.2017.58},
  \url{https://doi.org/10.1109/WACV.2017.58}

\bibitem{DBLP:journals/corr/abs-1708-07120}
Smith, L.N., Topin, N.: Super-convergence: Very fast training of residual
  networks using large learning rates. CoRR  \textbf{abs/1708.07120} (2017),
  \url{http://arxiv.org/abs/1708.07120}

\bibitem{DBLP:journals/jmlr/SrivastavaHKSS14}
Srivastava, N., Hinton, G.E., Krizhevsky, A., Sutskever, I., Salakhutdinov, R.:
  Dropout: a simple way to prevent neural networks from overfitting. J. Mach.
  Learn. Res.  \textbf{15}(1),  1929--1958 (2014),
  \url{http://dl.acm.org/citation.cfm?id=2670313}

\bibitem{DBLP:conf/cade/Tammet19}
Tammet, T.: {GKC:} {A} reasoning system for large knowledge bases. In: Fontaine
   \cite{DBLP:conf/cade/2019}, pp. 538--549.
  \doi{10.1007/978-3-030-29436-6\_32},
  \url{https://doi.org/10.1007/978-3-030-29436-6\_32}

\bibitem{logits_in_mireks_network}
Urban, J.: personal communication

\bibitem{DBLP:conf/nips/VaswaniSPUJGKP17}
Vaswani, A., Shazeer, N., Parmar, N., Uszkoreit, J., Jones, L., Gomez, A.N.,
  Kaiser, L., Polosukhin, I.: Attention is all you need. In: Guyon, I., von
  Luxburg, U., Bengio, S., Wallach, H.M., Fergus, R., Vishwanathan, S.V.N.,
  Garnett, R. (eds.) Advances in Neural Information Processing Systems 30:
  Annual Conference on Neural Information Processing Systems 2017, December
  4-9, 2017, Long Beach, CA, {USA}. pp. 5998--6008 (2017),
  \url{https://proceedings.neurips.cc/paper/2017/hash/3f5ee243547dee91fbd053c1c4a845aa-Abstract.html}

\bibitem{DBLP:conf/cav/Voronkov14}
Voronkov, A.: {AVATAR:} the architecture for first-order theorem provers. In:
  Biere, A., Bloem, R. (eds.) Computer Aided Verification - 26th International
  Conference, {CAV} 2014, Held as Part of the Vienna Summer of Logic, {VSL}
  2014, Vienna, Austria, July 18-22, 2014. Proceedings. Lecture Notes in
  Computer Science, vol.~8559, pp. 696--710. Springer (2014).
  \doi{10.1007/978-3-319-08867-9\_46},
  \url{https://doi.org/10.1007/978-3-319-08867-9\_46}

\bibitem{DBLP:conf/cade/WeidenbachDFKSW09}
Weidenbach, C., Dimova, D., Fietzke, A., Kumar, R., Suda, M., Wischnewski, P.:
  {SPASS} version 3.5. In: Schmidt, R.A. (ed.) Automated Deduction - CADE-22,
  22nd International Conference on Automated Deduction, Montreal, Canada,
  August 2-7, 2009. Proceedings. Lecture Notes in Computer Science, vol.~5663,
  pp. 140--145. Springer (2009). \doi{10.1007/978-3-642-02959-2\_10},
  \url{https://doi.org/10.1007/978-3-642-02959-2\_10}

\end{thebibliography}

\end{document}